\documentclass{article}
\usepackage{spconf,amsmath,graphicx, booktabs, multirow, tabularx,makecell}

\usepackage{enumitem}
\setlist{nosep, leftmargin=14pt}

\usepackage{mwe} 

\title{SAM-driven weakly supervised nodule segmentation with uncertainty-aware cross teaching}
\name{Xingyue Zhao$^{1}$, Peiqi Li$^{1}$, Xiangde Luo$^{2,3}$, Meng Yang$^{5}$, Shi Chang$^{4*}$, Zhongyu Li$^{1}$\sthanks{Correspondence: zhongyuli@xjtu.edu.cn, changshi@csu.edu.cn}}
\address{$^{1}$School of Software Engineering, Xi’an Jiaotong University, Xi’an,
China \\
$^{2}$University of Electronic Science and Technology of China, Chengdu, China \\
$^{3}$Shanghai Artificial Intelligence
Laboratory, Shanghai 200030, China \\
$^{4}$Department of General Surgery, Xiangya Hospital, Central South University, China \\
$^{5}$Hunan Frontline Medical Technology Co., Ltd
}

\begin{document}
%
\maketitle

\begin{abstract}
Automated nodule segmentation is essential for computer-assisted diagnosis in ultrasound images. Nevertheless, most existing methods depend on precise pixel-level annotations by medical professionals, a process that is both costly and labor-intensive. Recently, segmentation foundation models like SAM have shown impressive generalizability on natural images, suggesting their potential as pseudo-labelers. However, accurate prompts remain crucial for their success in medical images. In this work, we devise a novel weakly supervised framework that effectively utilizes the segmentation foundation model to generate pseudo-labels from aspect ration annotations for automatic nodule segmentation. Specifically, we develop three types of bounding box prompts based on scalable shape priors, followed by an adaptive pseudo-label selection module to fully exploit the prediction capabilities of the foundation model for nodules. We also present a SAM-driven uncertainty-aware cross-teaching strategy. This approach integrates SAM-based uncertainty estimation and label-space perturbations into cross-teaching to mitigate the impact of pseudo-label inaccuracies on model training. Extensive experiments on two clinically collected ultrasound datasets demonstrate the superior performance of our proposed method.

\end{abstract}
%
\section{Introduction}
\label{sec:intro}
Deep learning has significantly advanced automatic medical image segmentation, key for computer-assisted diagnosis. Most of the existing methods need extensive, precise pixel-level annotations, which is a laborious process, especially for ultrasound images with their lower resolution and irregular lesions.
\begin{figure}[htbp]
	\centering
	\setlength{\abovecaptionskip}{-0.2cm}  
	\setlength{\belowcaptionskip}{-1cm}
	\centerline{\includegraphics[width=7.0cm]{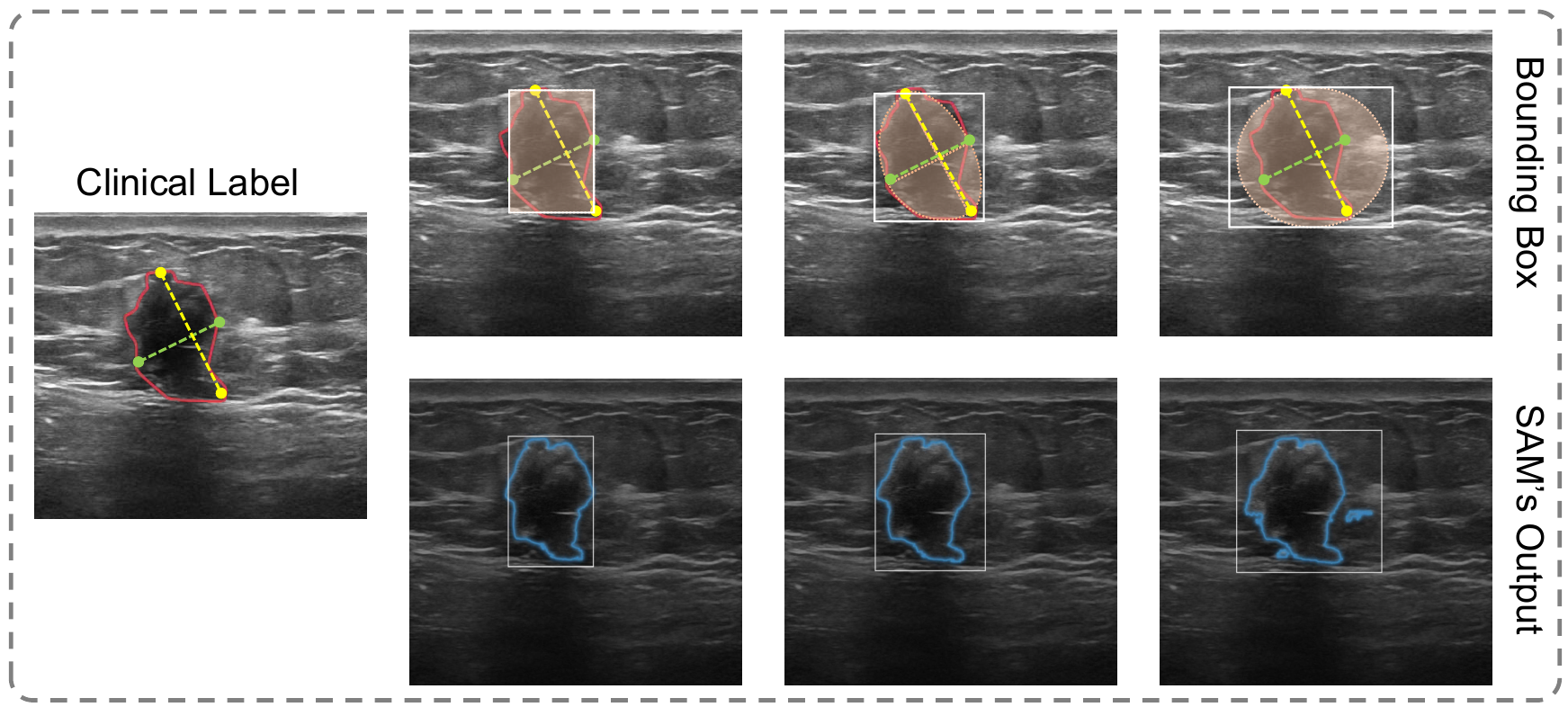}}
	\caption{Examples for aspect ratio annotation and prediction results of segment anything model in different bounding box prompts setting.}\medskip
	\vspace{-0.7cm}
	\label{intro}
\end{figure}
Recently, researchers have explored weakly supervised learning to reduce the need for extensive data annotation. Grandvalet et al.~\cite{grandvalet2004semi} proposed a method that diminishes uncertainty in unannotated data, allowing the model to produce high-confidence predictions. Javanmardi et al.~\cite{javanmardi2016unsupervised} emphasized promoting spatial smoothness by reducing pixel variation among neighboring regions. Kim et al.~\cite{kim2019mumford} introduced a novel loss function that balances segmentation smoothness with data fidelity, thereby promoting consistent segmentations. Additionally, several researchers have explored strategies to train models using sparse annotations, including image-level annotations \cite{papandreou2015weakly}, scribbles \cite{luo2022scribble,lin2016scribblesup}, bounding boxes \cite{dai2015boxsup}, and point annotations \cite{zhai2023pa}. For example, in the realm of tumor segmentation, some researchers have tried to train models using RECIST (Response Evaluation Criteria in Solid Tumors) annotations \cite{zhou2023recist,zhou2023recist2,wang2022recistsup,cai2018accurate}.

Despite recent advancements, current methods still exhibit significant challenges. One primary issue is the insufficient detail provided by weak annotations, especially concerning the location and shape of lesions. The training process is frequently misled by this limitation, and the diverse shapes and textures of nodules in ultrasound images further intensify this issue. Furthermore, while weak annotations are more cost-effective, they still require medical professionals to manually annotate the training set. Many existing methods depend on unsupervised segmentation algorithms, such as GrabCut \cite{rother2004grabcut}, to generate pseudo-labels. Often, these approaches \cite{wang2022recistsup, cai2018accurate} necessitate the use of custom trimaps and iterative updates, leading to unstable performance and extended computational times, particularly with low-resolution ultrasound images. Some researchers  \cite{zhou2023recist,zhou2023recist2} have tried to directly construct pseudo-labels from weak annotations. However, these static pseudo-labels are not adaptive enough to cater to the diverse shapes of lesions.
\begin{figure*}[htbp]
	\centering
	\vspace{-0.8cm}
	\setlength{\abovecaptionskip}{-0.2cm}   
	\setlength{\belowcaptionskip}{-1cm}   
    \includegraphics[width=0.9\linewidth]{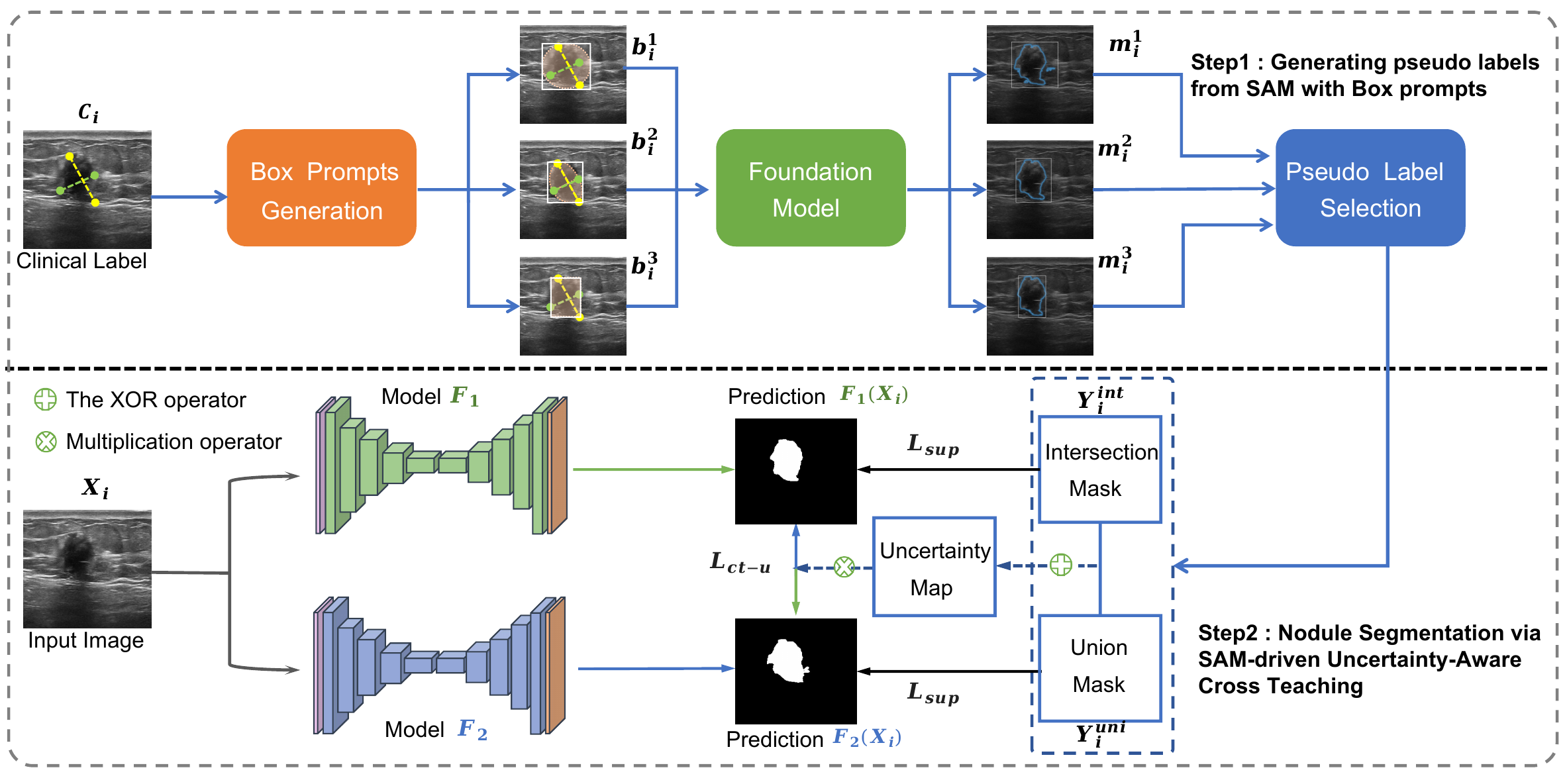}
	\vspace{-0.3cm}
	\caption{Overview of our proposed framework. The framework includes two stages. In the initial stage, three distinct sets of prior bounding boxes are created based on clinical annotations and fed as prompts into the SAM, generating pseudo labels. In the subsequent stage, a selection is made from these pseudo labels. A strategy named SAM-driven Uncertainty-Aware Cross Teaching is then introduced, leveraging these selected pseudo labels for nodule segmentation.}\medskip
	\vspace{-0.5cm}
	\label{model}
\end{figure*}

Distinct from other forms of sparse annotation, aspect ratio annotations (clinical annotations) from clinical ultrasound provide valuable nodule shape and location information (see Fig.~\ref{intro}). Readily accessible in hospital PACS, they offer a convenient source of weak annotations for nodule segmentation. The main challenge is to devise a method that can use these annotations to generate useful pseudo-labels for training models efficiently. In recent developments, foundational models like SAM \cite{kirillov2023segment}, trained on extensive datasets, aim to reduce reliance on detailed manual annotations for image segmentation tasks. While segmentation foundational models excel with natural images, they struggle with medical images, particularly those with low contrast like ultrasound images. Using bounding boxes as prompts can improve segmentation performance, but results heavily depend on the bounding boxes' precision.

In this study, prompted by the identified challenges, we introduce a novel framework for automated nodule segmentation utilizing aspect ratio annotations. Specifically, we leverage SAM for pseudo-label generation. However, there is a particular concern that bounding boxes generated directly from clinical annotations may not fully encompass nodules, especially since nodules often have a convex shape \cite{alexander2004thyroid} (see Fig.~\ref{intro}). To overcome the limitations of bounding boxes, we propose a Box Prompts Generation (BPG) module that creates bounding box prompts with clinical shape priors. These prompts are then used in conjunction with SAM to generate pseudo-labels. Subsequently, we designed a pseudo-label selection module that conducts pixel-level selection from the three potential pseudo-labels produced by SAM, resulting in two final pseudo-labels. Additionally, we have developed an SAM-driven uncertainty estimation method that utilizes the inconsistencies in SAM’s outputs for the BPG-generated bounding boxes. Ultimately, we employ a SAM-driven Uncertainty-Aware Cross Teaching strategy for model training. This involves two models trained separately with the two final pseudo-labels engaging in cross-supervised training under the guidance of uncertainty maps. Extensive experiments on two clinical ultrasound datasets, thyroid and breast, reveal that our framework outperforms existing approaches in clinical label settings.

\section{Method}
\label{sec:format}

\subsection{Overview}
\label{ssec:m1}
The overall framework of our proposed method is illustrated in Figure 2, comprising two primary steps. Initially, the Box Prompts Generation module is employed to create bounding box prompts with shape prior, which serve to enhance SAM’s segmentation capabilities. Subsequently, a pixel-level pseudo-label selection module is utilized to generate two pseudo-labels. In the second step, these pseudo-labels from the prior stage are used to train two fully automated segmentation models. We introduced a new uncertainty estimation method and devised a SAM-driven Uncertainty-Aware Cross Teaching strategy to achieve improved segmentation results.

\subsection{Generating pseudo labels from SAM}
\label{ssec:m2}
\subsubsection{Generating bounding box prompts}
\label{sssec:gbbp}
We define the dataset with clinical labels as $\mathcal{D}$, with each labeled data pair represented by $(\mathbf{X}, \mathbf{C}) \in \mathcal{D}$, where $\mathbf{X}$ is the raw image, and $\mathbf{C}$ is the corresponding clinical label. A precise prompt is instrumental in enhancing the segmentation performance of the Segment Anything Model (SAM). For each clinical label $\mathbf{c}_{i}$, we initially generate a tight bounding box as a prompt, denoted by $\mathbf{b}^{1}$. Given that nodules often exhibit a convex shape, a tight bounding box may not fully capture the lesion's extent. To address this, we propose the generation of bounding boxes with shape priors. Specifically, we construct an approximate ellipse by connecting curved lines between each pair of clinically annotated adjacent endpoints and derive a minimum enclosing circle based on the clinical annotations. Subsequently, we generate two bounding boxes based on the approximate ellipse and the minimum enclosing circle, denoted as $\mathbf{b}^{2}$ and $\mathbf{b}^{3}$, respectively. 

\subsubsection{Pseudo label selection}
\label{sssec:pls}
For the $i$-th image in our dataset, three bounding box prompts have been generated, denoted as $\mathbf{B}_i=\left\{b^{1}_i, b^{2}_i, b^{3}_i\right\}$. Subsequently, each bounding box prompt is used to guide SAM in generating a corresponding pseudo-label. This process yields three pseudo-labels, expressed as $\mathbf{m}^{1}_i$, $\mathbf{m}^{2}_i$ and $\mathbf{m}^{3}_i$. Under the guidance of shape-prior bounding box prompts, we derive the minimum and maximum predicted masks from SAM's outputs by calculating the intersection and union of $\mathbf{m}^{1}_i$, $\mathbf{m}^{2}_i$ and $\mathbf{m}^{3}_i$. Formally, we define the minimum predicted mask as $\mathbf{Y}^{int}_i$ and the maximum predicted mask as $\mathbf{Y}^{uni}_i$. These can be computed respectively as follows:
\begin{equation}
\begin{aligned}
\mathbf{Y}^{int}_i = \mathbf{m}^{1}_i \cap \mathbf{m}^{2}_i\cap \mathbf{m}^{3}_i, \\
\mathbf{Y}^{uni}_i = \mathbf{m}^{1}_i \cup \mathbf{m}^{2}_i\cup \mathbf{m}^{3}_i,
\end{aligned}
\end{equation}
where $\cap$ denotes the intersection operation and $\cup$ signifies the union operation.

\subsection{Learning from pseudo labels}
\label{ssec:m3}
\subsubsection{Uncertainty Estimation}
\label{sssec:ueos}
SAM’s performance is sensitive to the position and size of bounding box. Despite the incorporation of shape priors into our bounding box generation, ensuring absolute precision remains a challenge. To address this, we propose a novel SAM-driven Uncertainty Estimation approach to inform and enhance the model training process. Specifically, regions where predictions $\mathbf{m}^{1}_i$, $\mathbf{m}^{2}_i$ and $\mathbf{m}^{3}_i$ concur are considered certain, while discrepancies among these predictions denote areas of uncertainty. The uncertainty map of the $i$-th image, denoted as $\mathbf{U}_i$, can be computed as follows:
\begin{equation}
\mathbf{U}_{i}=\mathbf{Y}_{i}^{int} \oplus \mathbf{Y}_{i}^{uni},
\end{equation}
where $\oplus$ represents the pixel-wise XOR operator to indicate the inconsistency region between $\mathbf{Y}_{i}^{int}$ and $\mathbf{Y}_{i}^{uni}$.

\subsubsection{SAM-driven Uncertainty-Aware Cross Teaching}
\label{sssec:suct}
The proposed methodology incorporates a dual-model architecture for segmentation, where $\mathcal{F}_{1}$ and $\mathcal{F}_{2}$ represent the two segmentation models. Perturbations in label-space are introduced to enhance model robustness. Specifically, $\mathcal{F}_{1}$ is trained using the intersection of pseudo-labels $\mathbf{Y}^{int}_i$, while $\mathcal{F}_{2}$ is trained using the union of pseudo-labels $\mathbf{Y}^{uni}_i$. The supervised loss function is a composite of two commonly employed loss functions in segmentation tasks, which are defined as follows:
\begin{equation}
\begin{aligned}
\mathcal{L}_{sup} = &\mathcal{L}_{ce}(\mathcal{F}_{1}(\mathbf{X}_{i}), \mathbf{Y}_{i}^{int}) + \mathcal{L}_{Dice}(\mathcal{F}_{1}(\mathbf{X}_{i}), \mathbf{Y}_{i}^{int}) \\
+&\mathcal{L}_{ce}(\mathcal{F}_{2}(\mathbf{X}_{i}), \mathbf{Y}_{i}^{uni}) + \mathcal{L}_{Dice}(\mathcal{F}_{2}(\mathbf{X}_{i}), \mathbf{Y}_{i}^{uni}).
\end{aligned}
\end{equation}
Owing to the distinct output-level properties resulting from label-space perturbations, we introduce the cross teaching between $\mathcal{F}_{1}$ and $\mathcal{F}_{2}$. Specifically, the prediction of a network is used as the pseudo label to supervise the other network. The pseudo labels can be computed as follows:
\begin{equation}
\begin{aligned}
\mathbf{Y}_i^{pl1}=\operatorname{argmax} (\mathcal{F}_{1}(\mathbf{X}_i)), \\
\mathbf{Y}_i^{pl2}=\operatorname{argmax} (\mathcal{F}_{2}(\mathbf{X}_i)).
\end{aligned}
\end{equation}
Given that label-space perturbations originate from SAM, we incorporate the SAM-based uncertainty estimation into the cross teaching process. Specifically, cross teaching is conducted solely within regions of uncertainty. Then, the SAM-driven cross teaching loss is defined as:
\begin{equation}
\begin{aligned}
\mathcal{L}_{ct-u}=\mathcal{L}_{ce}(\mathcal{F}_{1}(\mathbf{X}_i),\mathbf{Y}_i^{pl2},\mathbf{U}_i) \\
+\mathcal{L}_{ce}(\mathcal{F}_{2}(\mathbf{X}_i),\mathbf{Y}_i^{pl1},\mathbf{U}_i),
\end{aligned}
\end{equation}
where $\mathcal{L}_{ce}$ is cross entropy loss function. Finally, the overall objective function $\mathcal{L}_{total}$ can be defined as:
\begin{equation}
\mathcal{L}_{{total}}=\mathcal{L}_{{sup}}+\lambda\mathcal{L}_{ct-u}.
\label{tab:tol}
\end{equation}
where $\lambda$ is the weight factor. The determination of $\lambda$ was based on experimentation.

\section{Experiment and Results}
\label{sec:pagestyle}

\subsection{Datasets}
\label{ssec:subhead3}
We conducted the evaluation of the proposed method on two clinical ultrasound datasets for thyroid and breast nodules. The Thyroid Ultrasound Dataset consists of 422 left and 422 right thyroid ultrasound images, and the Breast Ultrasound Dataset comprises 604 left and 151 right breast ultrasound images. All the nodules were manually segmented by two radiologists, each with over a decade of experience. For further validation, these initial annotations were also reviewed by a third radiologist with over twenty years of experience. The dataset was randomly split into training and test subsets, following a ratio of approximately 4:1. During the training phase, we ensured that each image in the training set was accompanied by its respective clinical aspect ratio annotation.
\begin{table}[ht]
\vspace{-0.7cm}
\setlength{\tabcolsep}{1mm}
\centering
\caption{Comparison results of different methods on the two test datasets.}
\vspace{0.3cm}
\label{tab:table1}
\resizebox{1.0\columnwidth}{!}{
\begin{tabular}{ccccccc} 
    \specialrule{.8pt}{0pt}{2pt}
    \multirow{2}{*}{Method} &
    \multicolumn{2}{c}{Thyroid Ultrasound}  && \multicolumn{2}{c}{Breast Ultrasound} \\ 
    \cline{2-3} \cline{5-6} 
    &  DSC(\%)&HD95(pixel)&&DSC(\%)&HD95(pixel) \\
    \specialrule{.4pt}{2pt}{0pt}
    EM~\cite{grandvalet2004semi} & 71.7$\pm$23.6 & 35.6$\pm$68.4 && 72.4$\pm$21.0 & 62.5$\pm$43.3  \\
    TV~\cite{javanmardi2016unsupervised} & 70.7$\pm$23.5 & 53.4$\pm$95.8 && 73.4$\pm$21.0 & 59.3$\pm$43.0  \\
    Mumford-Shah~\cite{kim2019mumford}  & 71.7$\pm$22.6 & 36.3$\pm$71.4 && 72.3$\pm$24.1 & 57.7$\pm$45.0  \\
    GatedCRF~\cite{obukhov2019gated}  & 68.6$\pm$25.2 & 39.1$\pm$74.6 && 73.2$\pm$20.3 & 57.7$\pm$39.6  \\
    WSSS~\cite{cai2018accurate}  & 73.3$\pm$21.3 & 37.2$\pm$67.1 && 74.7$\pm$19.5 & 59.8$\pm$41.9  \\
    RECISTSup~\cite{wang2022recistsup}  & 72.5$\pm$21.9 & 37.5$\pm$72.9 && 72.5$\pm$20.8 & 61.6$\pm$41.1  \\
    CoTraining~\cite{zhou2023recist}  & 73.4$\pm$23.7 & 39.2$\pm$74.0 && 68.3$\pm$22.0 & 58.9$\pm$37.5  \\
    CoTeaching~\cite{han2018co} & 71.2$\pm$24.0 & 38.7$\pm$72.5 && 72.7$\pm$21.8 & 53.4$\pm$37.3 \\
    TriNet~\cite{zhang2020robust} & 70.8$\pm$23.9 & 35.9$\pm$72.4 && 69.5$\pm$22.6 & 58.2$\pm$38.8 \\
    MTCL~\cite{xu2022anti} & 70.2$\pm$25.0 & 40.1$\pm$76.3 && 70.5$\pm$22.1 & 60.1$\pm$38.6 \\
    Ours & \textbf{73.5$\pm$22.2} &\textbf{32.8$\pm$71.5} && \textbf{75.5$\pm$\textbf{19.8}}& \textbf{52.3$\pm$36.5} \\ \hline 
    GrabCut~\cite{rother2004grabcut} & 68.6$\pm$25.3 & 48.4$\pm$85.2 && 71$\pm$23.0 & 59.2$\pm$42.7  \\
    SAM~\cite{kirillov2023segment} & 72.1$\pm$23.5 & 42.0$\pm$78.7 && 74.6$\pm$20.0 & 59.2$\pm$42.4  \\
    Fully Supervised & 74.7$\pm$23.8 & 35.3$\pm$75.3 && 76.4$\pm$20.6 & 53.2$\pm$43.2  \\
    \specialrule{.8pt}{0pt}{2pt}
\end{tabular}
}
\vspace{-0.8cm}
\end{table}

\subsection{Implementation Details}
\label{ssec:subhead1}
All compared methods were implemented in PyTorch \cite{paszke2019pytorch} utilizing an NVIDIA RTX 3090 GPU. To maintain consistency and ensure fairness in our comparisons, we adopted the UNet \cite{unet2015MICCAI} as the common backbone network for all the methods involved in our experiments. All images were uniformly resized to dimensions of 256 × 256 pixels and each experiment underwent an identical data augmentation process including random rotation and flipping. The batch size was set to 24. The learning rate was initially set at 0.01, and the poly learning rate strategy \cite{luo2021efficient} was employed for adaptive fine-tuning of the learning rate. We evaluated performance using the Dice similarity coefficient (DSC) and 95th percentile Hausdorff Distance (HD95).

\subsection{Results}
\label{ssec:subhead2}
\noindent{\bf Comparison With State-of-the-Arts.}
To assess the performance of our framework, we compared it with several state-of-the-art weakly supervised approaches including GrabCut only~\cite{rother2004grabcut}, Entropy Minimization (EM)~\cite{grandvalet2004semi}, Total Variation (TV) loss~\cite{javanmardi2016unsupervised}, Mumford-Shah loss~\cite{kim2019mumford}, GatedCRF Loss~\cite{obukhov2019gated}, WSSS~\cite{cai2018accurate}, RECISTSup~\cite{wang2022recistsup} and CoTraining~\cite{zhou2023recist}. Additionally, as the pseudo-labels may contain noise, we evaluated our method alongside other popular approaches for learning from noisy annotations: 1) CoTeaching~\cite{han2018co}, 2) TriNet~\cite{zhang2020robust}, 3) MTCL~\cite{xu2022anti}. All compared methods, except for CoTraining \cite{zhou2023recist}, use GrabCut to generate pseudo-labels for training. The results of these methods are tabulated in Table \ref{tab:table1}.
\begin{table}[htbp]
\centering
\vspace{-0.2cm}
\caption{Ablation studies of our proposed framework. We compared our proposed method with different dual model supervision strategies on the Thyroid Ultrasound dataset.}
\vspace{0.4cm}
\label{tab:table2}
\begin{tabular}{l|cc}
\toprule 
Method                                                   & \multicolumn{1}{c}{DSC}  & HD95  \\ \hline
$\mathcal{F}_{single}$  &72.1$\pm$23.5 & 42.0$\pm$78.7     \\
Dual Model+CR~\cite{zhou2023recist,dolz2021teach} & 72.1$\pm$24.1 & 38.9$\pm$71.5    \\
Dual Model+CPS~\cite{chen2021semi} & 71.7$\pm$24.7 & 42.2$\pm$80.6   \\
Dual Model+DMPLS~\cite{luo2022scribble} & 73.1$\pm$23.0 & 37.9$\pm$73.2    \\
UAMT~\cite{yu2019uncertainty} & 71.7$\pm$24.2 & 40.9$\pm$81.9    \\ \hline
Ours  & \textbf{73.5$\pm$22.2} & \textbf{32.8$\pm$71.5} \\ \bottomrule 
\end{tabular}
\vspace{-0.4cm}
\end{table}

\noindent{\bf Ablation Study.}
Given the utilization of two networks in our framework, we compared the effects of various dual-model supervision methods with our proposed method. These approaches include: 1) Baseline (single model $\mathcal{F}_{single}$), 2) Consistency Regularization (CR)~\cite{zhou2023recist,dolz2021teach}, 3) Cross Pseudo Supervision (CPS)~\cite{chen2021semi}, 4) Dynamically Mixed Pseudo Labels Supervision (DMPLS)~\cite{luo2022scribble}, 5) SAM-driven Uncertainty-Aware Cross Teaching (Ours). Additionally, we also compared our SAM-driven uncertainty estimation with traditional method using Monte Carlo simulation (UAMT)~\cite{yu2019uncertainty}. The quantitative evaluation results are presented in Table \ref{tab:table2}.

\begin{figure}[htbp]
	\centering
	\vspace{-0.2cm}
	\setlength{\abovecaptionskip}{-0.2cm}   
	\setlength{\belowcaptionskip}{-1cm}   
	\centerline{\includegraphics[width=8.0cm]{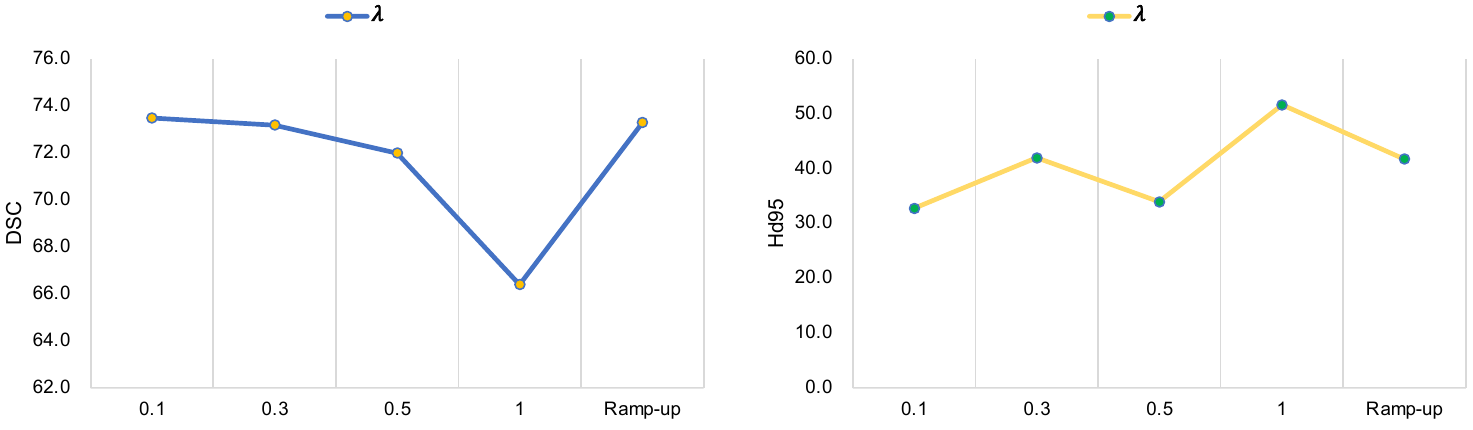}}
	\caption{The impact of varying $\lambda$ on the segmentation performance of the proposed framework, assessed on the Thyroid Ultrasound dataset using Dice similarity coefficient (DSC) and 95th percentile Hausdorff Distance (HD95).}\medskip
	\vspace{-0.7cm}
	\label{fig:para}
\end{figure}

\noindent{\bf Hyper-parameters Experiments Results.}
The proposed framework has a hyper-parameter $\lambda$. we conducted experiments to investigate the impact of different values of the hyper-parameter $\lambda$. Specifically, we evaluated our method across different settings of $\lambda \in \{0.1,0.3,0.5,1\}$. Additionally, we also explored a dynamic approach to setting $\lambda$ by employing a Gaussian warming-up function dependent on time \cite{yu2019uncertainty}. As shown in Fig.~\ref{fig:para}, the best performance was achieved when $\lambda$ was set to 0.1.

\section{Conclusion}
\label{sec:typestyle}

In this paper, we propose a novel weakly supervised framework for nodule segmentation using clinical annotations, which effectively employs the zero-shot segmentation capabilities of the foundational segmentation model with a shape-prior box prompts generation module and a SAM-driven uncertainty-aware cross-teaching strategy. Extensive experiments conducted demonstrate that our framework achieves state-of-the-art performance.

\vfill
\pagebreak

\section{Acknowledgments}
\label{sec:acknowledgments}
This work is supported by the Key Research and Development Program of Shaanxi Province under Grant. 2021GXLH-Z-097.


\bibliographystyle{IEEEbib}
\bibliography{ref}

\end{document}